\documentclass[conference]{IEEEtran}
\IEEEoverridecommandlockouts
\usepackage{comment}
\usepackage{amsmath}
\usepackage{algorithm}
\usepackage{booktabs}
\usepackage{multirow}
\usepackage{siunitx}
\usepackage{array}
\usepackage[table]{xcolor}
\usepackage{adjustbox}   % for max width=\textwidth

\usepackage{cite}
\usepackage{amsmath,amssymb,amsfonts}
\usepackage{algorithmic}
\usepackage{graphicx}
\usepackage{textcomp}
\usepackage{xcolor}
\def\BibTeX{{\rm B\kern-.05em{\sc i\kern-.025em b}\kern-.08em
    T\kern-.1667em\lower.7ex\hbox{E}\kern-.125emX}}

\begin{document}

\title{SPAR-K: Scheduled Periodic Alternating Early Exit for Spoken Language Models}

\author{\IEEEauthorblockN{Hsiao-Ying Huang}
\IEEEauthorblockA{\textit{National Taiwan University}\\
Taipei, Taiwan \\
monicahuangml1@gmail.com}
\and
\IEEEauthorblockN{Cheng-Han Chiang}
\IEEEauthorblockA{\textit{National Taiwan University}\\
Taipei, Taiwan \\
dcml0714@gmail.com}
\and
\IEEEauthorblockN{Hung-yi Lee}
\IEEEauthorblockA{\textit{National Taiwan University}\\
Taipei, Taiwan \\
tlkagkb93901106@gmail.com}
}% \author{Anonymous Submission}

\maketitle

\begin{abstract}
Interleaved spoken language models (SLMs) alternately generate text and speech tokens, but decoding at full transformer depth for every step becomes costly, especially due to long speech sequences. 
We propose SPAR-K, a modality-aware early exit framework designed to accelerate interleaved SLM inference while preserving perceptual quality. 
SPAR-K introduces a speech alternating-depth schedule: most speech positions exit at a fixed intermediate layer, while periodic full-depth "refresh" steps mitigate distribution shift due to early exit.
We evaluate our framework using Step-Audio-2-mini and GLM-4-Voice across four datasets spanning reasoning, factual QA, and dialogue tasks, measuring performance in terms of ASR transcription accuracy and perceptual quality.
Experimental results demonstrate that SPAR-K largely preserves question-answering accuracy with a maximum accuracy drop of 0.82\% while reducing average speech decoding depth by up to 11\% on Step-Audio-2-mini and 5\% on GLM-4-Voice, both with negligible changes in MOS and WER and no auxiliary computation overhead.
We further demonstrate that confidence-based early exit strategies, widely used in text LLMs, are suboptimal for SLMs, highlighting that the unique statistical nature of speech tokens necessitates a specialized early exit design. 
\end{abstract}


\section{Introduction}
Spoken language models (SLMs) aim to unify speech understanding and speech generation in a single autoregressive model, taking speech as input and producing speech output~\cite{arora2025landscape, cui2025recent, defossez2024moshi, ding2025kimi}.
Recent end-to-end systems demonstrate that large transformer backbones can be trained to support speech-to-speech conversation by modeling a unified token stream that includes both text and discrete speech units~\cite{zeng2024scaling,zhang2023speechgpt, cui2025recent, defossez2024moshi, ding2025kimi}. 
A particularly simple and increasingly prospective structure is the interleaved SLM, which generates speech output by alternatively generating text and speech tokens in a fixed ratio, enabling streaming synthesis and, importantly, allowing text tokens to provide explicit semantic guidance for subsequent speech tokens~\cite{yang2024interleaved, chou2023toward,zeng2024glm,wu2025step,li2025baichuan, chiang2025stitch}. 
The speech tokens will be synthesized to audio via an audio decoder~\cite{du2024cosyvoice2, du2024cosyvoice, kong2020hifi}.

Despite their impressive capabilities, modern SLMs are computationally expensive at inference time: they inherit the depth and width of large language models, and must additionally decode long sequences of speech units, making real-time deployment challenging~\cite{arora2025landscape, defossez2024moshi, li2025baichuan}.
In text-only LLM inference, a large body of work reduces decoding cost via adaptive compute, including \textit{early exit} strategies that dynamically allocate per-token "thinking depth" based on confidence signals~\cite{schuster2022confident, bae2023fast, valade2024accelerating, zheng2025hybrid, miao2024efficient, elhoushi2024layerskip}.
However, in our later experiment, we will show that directly transferring such early exit policies to interleaved SLMs may not always work, showing that text tokens and speech tokens are inherently different and cannot early exit in the same way.

We conduct an experiment to show that \textbf{text and speech tokens behave differently} for interleaved SLMs.
Given a speech input and the complete sequence of text-speech output tokens generated using standard decoding, for each generation step, we can extract the hidden representation from an intermediate layer and predict a token from this distribution using a trained layer-specific LM head (Sec.~\ref{ssec:tuned_lens}).
Note this experiment, we do \textit{not} use the token prediction from intermediate layers as the input for the model in the next generation step.
Interestingly, although intermediate-layer speech tokens often differ from the final-layer predictions, the synthesized audio still sounds similar. 
In contrast, the text tokens extracted from intermediate layers cannot form a coherent sentence. 
This suggests that computing full transformer depth at every speech position may be unnecessary, which motivates our early exit design on speech steps.

\begin{figure}[tbp]
  % \centering
  % \vspace{-10pt}
  \includegraphics[width=0.52\textwidth]{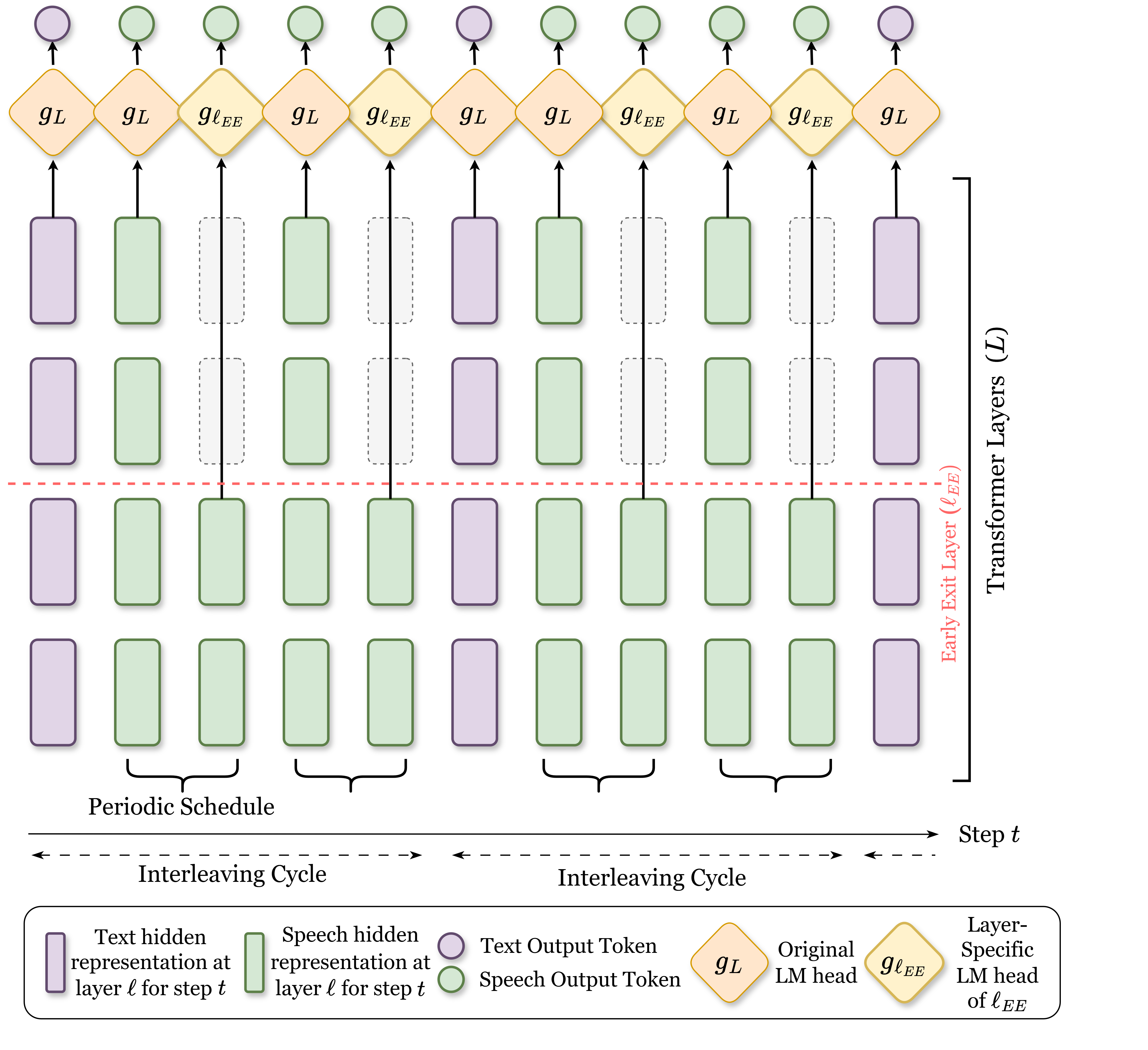}
  
  \caption{Overview of our method -- SPAR-K Early Exit Scheduler (take Step-Audio-2-mini for example)}
  \label{fig:methodology_overview}
  \vspace{-5pt}
\end{figure}

In this work, we propose a speech-specialized \textbf{scheduled periodic alternating early exit (SPAR-K)} framework for interleaved SLMs (shown in Fig.~\ref{fig:methodology_overview}). 
The core of the \textbf{SPAR-K} framework is a $\mathrm{K}$ scheduling strategy that selectively executes early exits for specific speech positions within the interleaving sequence. 
By interleaving early exit positions with full-depth inference steps, we provide the model with periodic "refreshes" that prevent the severe distribution shift from full-depth decoding.
Empirically, our approach preserves semantic perception quality on diverse tasks, including dialogue and factual QA, while reducing average decoding depth for speech tokens.
Contrary to the unstable transfer of the common confidence-based early exit paradigms (Sec.~\ref{subsection: Compared Methods}, Tab.~\ref{tab:main_results_grouped}: S3, G3), our schedule-based policy consistently provides a practical efficiency--quality trade-off without incurring additional per-step confidence computation on speech positions. 
Our main contributions are:
\begin{itemize}
    \item We are the first paper to explore early exit in interleaved SLMs. 
    We propose SPAR-K, an early exit policy that improves decoding efficiency without any auxiliary computation overhead. %without any per-step auxiliary computation at inference time
    \item Using two interleaved SLM backbones on four datasets, SPAR-K consistently reduces the computation by 5\% to 11\% while maintaining the semantic and perceptual quality of the speech response.
    \item We provide empirical evidence that the distinct nature of speech and text tokens in interleaved SLMs leads to distinct early exit policy.
\end{itemize}

\section{Related Work} %: Early Exit}

\subsection{Early Exit}
\label{ssec:ee_in_nlp}
Standard autoregressive decoding generates each token by passing through the full stack of transformer layers in the language model (LM). 
Early exit methods~\cite{bajpai2025survey, chen2023ee} accelerate generation by producing the next-token prediction at an intermediate (non-last) layer, and use this next token as the input to the model in the next generation step.
By skipping the remaining layers, the computation is saved. %skipping the remaining layers and saving their computation. 
In text-only LLMs, prior work falls
into two broad families: (1) \textbf{Training-enabled early exit} modifies pretraining or fine-tuning so that shallower layers produce usable predictions~\cite{elhoushi2024layerskip,schuster2022confident, elbayad2019depth}; (2) \textbf{Confidence-adaptive early exit} dynamically selects a per-token depth and exits once an intermediate prediction is sufficiently confident under some uncertainty indicator~\cite{schuster2022confident, bae2023fast,valade2024accelerating,zheng2025hybrid, miao2024efficient,elhoushi2024layerskip, chen2023accelerating}.

In speech systems, early exit has so far been explored only on \textit{discriminative} encoders for automatic speech recognition (ASR) and self-supervised representation learning~\cite{yoon2022hubert,wright2024training,lasbordes2025splitformer,lin2024daisy}. %, along three distinct axes. 
% HuBERT-EE~\cite{yoon2022hubert} attaches CTC branches to intermediate layers of a fine-tuned HuBERT and exits per utterance via CTC confidence or frame entropy. 
% \cite{wright2024training} study EE training dynamics, showing that jointly training all exits from scratch outperforms fine-tuning pretrained backbones, and propose a sentence-level N-best confidence as an alternative exit metric.
% Splitformer~\cite{lasbordes2025splitformer} is an architectural variant that inserts parallel downsampled branches inspired by Zipformer to improve performance at every exit, without modifying the exit-decision rule. 
% DAISY~\cite{lin2024daisy} trains exit branches with a self-supervised loss rather than a task-specific objective, yielding task-agnostic, data-adaptive exits that use deeper layers on noisier inputs. 
Despite their minor structural or training technique differences, all prior speech early exit methods target discriminative models with frame-level CTC supervision and confidence-thresholded exit. %over a fixed vocabulary, where confidence-thresholded exit is a natural fit. 
To our knowledge, no prior work explores early exit in large SLMs with autoregressive speech-token generation. %, where this assumption no longer holds.

\begin{comment}
Standard decoding generates each token by using the full stack of transformer layers in the language model (LM).
In text-only LLMs, early exit methods~\cite{bajpai2025survey, chen2023ee} accelerate autoregressive generation by optionally using an intermediate (non-last) transformer layer to predict the next token, and use this next token as the input to the model in the next generation step.
Since \textit{exit} the current generation step early by skipping the last few transformer layers, the computation of those layers is saved.
Summarized from previous works in NLP, early exit can be done using (1) \textbf{Training-enabled early exit}, which modifies the pretraining or fine-tuning process so that shallower layers produce usable predictions~\cite{elhoushi2024layerskip, schuster2022confident, elbayad2019depth}. 
(2) \textbf{Confidence-adaptive early exit} dynamically selects a per-token depth and exits once intermediate predictions become sufficiently confident according to uncertainty indicators~\cite{schuster2022confident, bae2023fast, valade2024accelerating, zheng2025hybrid, miao2024efficient, elhoushi2024layerskip, leviathan2023fast, chen2023accelerating}.
In speech systems, most related early exit works are \cite{yoon2022hubert,wright2024training,lasbordes2025splitformer,lin2024daisy}, which only implement on small automatic speech recognition (ASR) models using depth-wise confidence measurement.
However, currently, no prior work explores early exit in large SLMs with LM transformer backbones. %, which inherit the depth and width of LLMs.
\end{comment}

\subsection{Inference Acceleration in Spoken Language Models}
Prior work on accelerating autoregressive SLMs largely operates along two axes that are orthogonal to layer depth.
Along the \textbf{token-rate axis}, shortening the generated speech sequence directly reduces the number of forward passes per second of audio.
Hassid et al.~\cite{hassid2023textually} show that lower speech token rates even improve SLM lexical and syntactic modeling, and recent SLMs adopt lower frame rate codecs such as Mimi~\cite{defossez2024moshi}.
Along the \textbf{token axis}, multi-token prediction (MTP) augments the backbone with auxiliary heads that emit several future tokens per forward pass. 
SLAM-Omni~\cite{chen2025slam} introduces group modeling for accelerated speech token generation, and VocalNet~\cite{wang2025vocalnet} proposes a refined MTP design tailored to speech LLMs. 
Speech Speculative Decoding~\cite{lin2025accelerating} adapts speculative decoding to autoregressive speech LMs, using a lightweight draft model with parallel verification by the frozen target. 
Both families require training: token-rate methods require pretraining the SLM with a lower-rate codec, and token-axis methods require training auxiliary heads or a draft model.
SPAR-K accelerates SLM inference along a complementary axis, layer depth, and is composable with the token-rate and token-axis methods above.

\subsection{Modality Characteristics in Interleaved SLMs} %Interleaving Mechanisms in Spoken Language Models}
% \blue{simply cite some papers discussing the property differences of text and speech tokens, either in same-vocab SLMs, conventional ASRs or other speech-aware LMs}
Interleaved SLMs~\cite{yang2024interleaved, chou2023toward,zeng2024glm,wu2025step,li2025baichuan, chiang2025stitch} generate text tokens and discrete speech tokens within a single autoregressive stream, but these two token types exhibit markedly different statistical properties. %codec 
Discrete speech tokens are typically high-rate and locally predictable: recent analyses of speech codec modeling observe strong short-range dependency, where next-token prediction is dominated by nearby acoustic context and acoustic redundancy naturally existing in speech also persists in discrete representations~\cite{liu2025speech, aylett2006language, zhang2023speechtokenizer}. % codec, neural codec language modeling and longer-range history often becomes redundant
The locality and redundancy suggest that computing full-depth inference at every speech position may be unnecessary, motivating our shallower decoding for speech steps. % full-depth reasoning
\section{Notations and Preliminary}

SLMs~\cite{cui2025recent, defossez2024moshi, ding2025kimi} take speech input $\mathbf{x}_{\rm in}$ and generate speech outputs.
Since directly generating the speech output is challenging, interleaved SLMs generate the speech output by alternately generating $N_{\rm text}$ text tokens and $N_{\rm speech}$ speech tokens~\cite{arora2025landscape,zeng2024glm,wu2025step}.
The text tokens correspond to the transcription of the speech tokens.
The SLM is a stack of $L$ layer transformer language model that autoregressively generates the next token conditioning on the input $\mathbf{x}_{\rm in}$ and previously generated output tokens $y_{i}$.
Let $h_{t}^{(\ell)}\in\mathbb{R}^{d}$ denotes the hidden representation at output step $t$ and layer $\ell\in\{1, \cdots, L\}$, where $d$ is the hidden dimension size.
At a generation step $t$, the SLM predicts the next token distribution $p_t(\cdot |\mathrm{x}, y_{<t})$ over the vocabulary space by a language model head $g_{L}$
\begin{equation}
    p_t(\cdot |\mathrm{x}, y_{<t})=\text{softmax}(g_{L}(h_t^{(\ell)}).
\end{equation}

Standard SLMs generate each token by using the full stack of transformer layers $\{1,\cdots,L\}$.
The next token distribution is always predicted by the last layer's hidden representation $h_{t}^{(L)}$.
In early exit, at a generation step $t$, we accelerate the inference speed by \textit{optionally} using the hidden representation $h_{t}^{(\ell_{EE})}$ at a layer $\ell_{EE}<L$ to predict the next token prediction.

\begin{table}[t]
% \centering
\caption{Oracle leave-at-fixed-layer study: MOS / Top-1 Agreement Ratio (Agree \%) vs.\ fixed exit layer $\ell$ in speech positions.}
\small
\setlength{\tabcolsep}{4.6pt} %4.6
\renewcommand{\arraystretch}{1.10}
\newcommand{\best}[1]{\textbf{#1}}
\begin{tabular}{
  l
  *{6}{S[table-format=1.3, table-number-alignment=center]}
}
\toprule
& \multicolumn{6}{c}{Fixed exit layer $\ell$} \\
\cmidrule(lr){2-7}
Metric & {16} & {21} & {26} & {31} & {36} & {40} \\
\midrule
MOS $\uparrow$ & 2.922 & 2.905 & 2.978 & 2.985 & 2.996 & 3.004 \\
\addlinespace
\multicolumn{1}{c}{Agree(\%) $\uparrow$} &
\multicolumn{1}{S[table-format=2.2, table-number-alignment=center]}{19.14} &
\multicolumn{1}{S[table-format=2.2, table-number-alignment=center]}{22.13} &
\multicolumn{1}{S[table-format=2.2, table-number-alignment=center]}{32.83} &
\multicolumn{1}{S[table-format=2.2, table-number-alignment=center]}{39.57} &
\multicolumn{1}{S[table-format=2.2, table-number-alignment=center]}{46.21} &
\multicolumn{1}{S[table-format=3.0, table-number-alignment=center]}{100} \\
\bottomrule
\end{tabular}
\vspace{-8pt}
\label{tab:layer_mos_horizontal}
\end{table}

\vspace{-7pt}
\section{Methodology}

We design a simple method to accelerate the generation speed of interleaved SLM by only early exiting the speech tokens for the interleaved SLM using a fixed period schedule.
We focus on interleaved SLMs since it is an important SLM paradigm adopted by many widely used open-sourced SLMs~\cite{zeng2024glm,wu2025step,li2025baichuan}
.

\subsection{Motivation}
\label{ssec:motivation}
Our method is motivated by two empirical observations about interleaved SLM generation. First, speech tokens dominate per-response compute: across our four evaluation datasets, they account for 79.5\% (Step-Audio-2) and 81.2\% (GLM-4-Voice) of all generated tokens (averaged over 2.5k responses per model), making speech the natural target for early exit. Second, the following preliminary experiment suggests that speech tokens remain perceptually viable when predicted from intermediate layers.

Given input $\mathbf{x}_{\rm in}$, we first use an interleaved SLM, GLM-4-Voice~\cite{zeng2024glm}, to generate the response, including the speech and text token \textit{without using early exit}.
Next, given a generation step $t$ and the already generated token $y_t$, we use a trained prediction head (introduced in Section~\ref{ssec:tuned_lens}) to predict a token from the hidden representation $h_{t}^{(\ell)}$ at layer $\ell$; denote the predicted token as $y_t^{(\ell)}$.
We collect all the speech tokens predicted at layer $\ell$ and synthesize them into speech.
We report the mean opinion score (MOS) evaluated by UTMOS-v2~\cite{baba2024t05} for $\ell \in \{16,21,26,31,36,40\}$ in Table~\ref{tab:layer_mos_horizontal}.
We additionally report the average agreement ratio.
Agreement between layer $\ell$ and $L$ is 1 if $y_t^{(\ell)}=y_t^{L}$; otherwise 0. 
We surprisingly found that the MOS is consistent across different $\ell$, even if the speech tokens predicted at layer $\ell$ may not match the prediction from the last layer.
This hints that using non-last layer speech token predictions may be feasible.
However, if we auto-regressively generate the next speech token by early exiting from a fixed $\ell<L$, we found that the speech quality deteriorates significantly (Table~\ref{tab:main_results_grouped}: S2, G2).
Motivated by the above observation, we design a SPAR-K, a method that periodically interleaves between early exit and full-depth generation for speech tokens.

\subsection{Method}
We propose a simple method for early exiting speech tokens in interleaved SLMs.
In a chunk of $N_{\rm speech}$ speech tokens, we use a fixed schedule with a period of $K$ token to periodically enable or disable early exit with a fixed pattern:
we early exit from layer $\ell_{EE}$ for $K-1$ positions and use full-depth decoding from layer $L$ at one position. 
To early exit from layer $\ell_{EE}$, we train a prediction head (Section~\ref{ssec:tuned_lens}) to predict a token distribution from the hidden representation at layer $\ell$.

\subsubsection{SPAR-K Early Exit Scheduler} 
\label{ssec:alternative_depth}
SPAR-K early exit schedule partitions a speech chunk of $N_{\rm speech}$ tokens into subgroups of size $K$\footnote{The last subgroup may be less than $K$} and use full-depth decoding at a fixed position while early exit at other positions in the subgroup.
We explore the two schedules with $K=2$ and one with $K=3$:
(i) \textbf{Even schedule}: Within a chunk of $N_{\rm{speech}}$ speech token, the early exit pattern starting from the first position is $\{L,\ell_{EE},L,\ell_{EE},\cdots\}$; this notation means that for the 1$^{\text{st}}$, 3$^{\text{rd}}$, etc., positions in the chunk of $N_{\rm{speech}}$ speech token, we use full-depth decoding, while the 2$^{\text{nd}}$, 4$^{\text{th}}$, etc., positions, we early exit from layer $\ell_{EE}$.
Figure~\ref{fig:methodology_overview} shows an example of this schedule.
(ii) \textbf{Odd schedule}: Within a chunk of $N_{\rm{speech}}$ speech token, the early exit pattern is $\{\ell_{EE},L,\ell_{EE},L,\cdots\}$. 
(iii) \textbf{Triple schedule}: The early exit pattern within a speech token chunk is $\{L, \ell_{EE},\ell_{EE},L,\ell_{EE},\ell_{EE},\cdots\}$. 

\subsubsection{Training the Layer-Specific LM Head}
\label{ssec:tuned_lens}
Since the original LM head $g_{L}(\cdots)$ is only trained to recognize the hidden representation from the last layer ($L$), we cannot directly use $g_{L}$ for the hidden representation of a different layer. 
 
For each transformer layer $\ell\in\{1,\ldots,L\}$, we learn a layer-specific LM head that projects the intermediate hidden state $h_t^{(\ell)}\in\mathbb{R}^d$ into vocabulary space~\cite{belrose2023eliciting}.
The resulting next-token distribution of intermediate layer is 
\begin{equation} 
p_\ell(\cdot \mid y_{<t})=\operatorname{softmax}\left(g_\ell(h_t^{(\ell)})\right).
\end{equation}
When training the layer-specific LM head, the SLM backbone is fixed and not trained.
The layer-specific LM head $g_{\ell}$ is trained to predict the token distribution of the last layer LM head $g_L$ using cross-entropy loss.
\begin{equation}
    \mathcal{L}^{(\ell)}= \mathbb{E}_{(y_{<t})\sim\mathcal{D}}\left[-\sum_{v\in\mathcal{V}} p_{L}(v\mid y_{<t}) \log p_\ell(v\mid y_{<t})\right].
\end{equation}

\subsubsection{Generating Missing KV-Cache by Deferred Parallel Decoding}
Early exit at layer $\ell_{EE}$ for position $t$ computes only layers 1 to $\ell_{EE}$, leaving $K^{(\ell)}_t, V^{(\ell)}_t$ uncomputed for every $\ell > \ell_{EE}$. 
Any later full-depth step $t'>t$ must attend to position $t$ at these upper layers, so the missing entries must be filled in before they are needed. 
Prior work resolves this either by approximate state copying~\cite{schuster2022confident} or by reactive synchronized parallel decoding~\cite{bae2023fast}.

SPAR-K adopts parallel decoding~\cite{bae2023fast} but triggers it on its own periodic full-depth steps rather than reactively, which makes the fill-in essentially free. % follows the parallel-decoding approach~\cite{bae2023fast} but exploits its own periodic full-depth steps to make the fill-in essentially free.
Instead of backfilling the upper layers at each early-exit step, we \textit{defer} them: we cache only $h^{(\ell_{EE})}_t$ and mark $t$ as pending.
At the next full-depth step $t'$, we co-process the pending early-exit positions in parallel with $t'$ through the upper layers $\ell_{EE}{+}1:L$ in a \textit{single} attention call under a length-$K$ causal mask, analogous to a mini-prefill restricted to uncomputed upper layer (Algorithm~\ref{alg:spark}). % on top of the existing cache (Algorithm~\ref{alg:spark}).
% This step plays two roles: it \textit{refreshes} position $t'$ at full depth to limit distribution shift, and it \textit{completes} the deferred KV of the pending positions.
Every full-depth step $t'$ therefore restores complete KV at all layers for all preceding positions, recovering the standard decoding invariant.
Note that even though the full-depth pass may predict a different token at $t$, we retain the early exit prediction as the emitted token. %$t$'s full-depth prediction could yield different tokens, we remain adopting $\ell_{EE}$'s predictions as emitted tokens.

Deferring the KV completion is what makes it cheap. 
GPU decoding at a small batch size is memory-bound: per-layer latency is dominated by loading that layer's weights from GPU memory (HBM) rather than by the tensor arithmetic on the tokens~\cite{pope2023efficiently}. 
Completing each pending position separately would traverse the upper layers once
per position, reloading their weights each time. %Backfilling at each early-exit step would instead traverse the upper layers twice, loading their weights both times. 
Processing the pending
positions in one batched pass (in parallel) over $\ell_{EE}{+}1:L$ instead loads each upper-layer weight matrix once, so the completion costs essentially the same as a single full-depth pass. %  and amortizes it across the batch , By fusing the pending positions into one pass over $\ell_{EE}{+}1:L$, SPAR-K loads each upper-layer weight matrix once and amortizes it across the fused tokens, so the catch-up costs essentially the same as a single full-depth pass. 
The efficiency of SPAR-K therefore has two sources: the early-exit steps are shallower, and the deferred completion adds negligible overhead on the full-depth steps.

\vspace{-3pt}
\begin{algorithm}[htbp]
\footnotesize
\caption{SPAR-K KV-Cache: Deferred Parallel Decoding}
\label{alg:spark}
\begin{algorithmic}[1]
\REQUIRE heads $g_L, g_{\ell_{EE}}$; early-exit layer $\ell_{EE}$; schedule $s_1,\dots,s_{N_{\text{speech}}}\in\{\textsc{ee},\textsc{full}\}$ for one speech chunk
\STATE $\mathcal{P}\gets\varnothing$ \COMMENT{positions with deferred upper KV}
\FOR{$t = 1$ to $N_{\text{speech}}$}
  \STATE run layers $1$ to $\ell_{EE}$; cache lower KV; store $h^{(\ell_{EE})}_t$
  \IF{$s_t = \textsc{ee}$}
     \STATE $y_t \gets \arg\max\ g_{\ell_{EE}}(h^{(\ell_{EE})}_t)$; $\mathcal{P}\gets\mathcal{P}\cup\{t\}$ \COMMENT{defer upper KV}
  \ELSE
     \STATE batch $\mathcal{P}\cup\{t\}$ through layers $\ell_{EE}{+}1$ to $L$ (length-$K$ mask) \COMMENT{full-depth refresh at $t$; fill pending upper-layer KV}
     \STATE $y_t \gets \arg\max\ g_{L}(h^{(L)}_t)$;\quad $\mathcal{P}\gets\varnothing$
  \ENDIF
\ENDFOR
\end{algorithmic}
\end{algorithm}

\vspace{-6pt}
\section{Experiments}
\subsection{Setup} 
\noindent\textbf{Models.}\quad 
We include two widely used interleaved SLMs in our experiment: Step-Audio-2-mini~\cite{wu2025step} and GLM-4-Voice~\cite{zeng2024glm}.
We will use Step-Audio-2 to refer to Step-Audio-2-mini for short.
Step-Audio-2 is a 28-layer model with a text-speech interleaved schedule of $N_{\rm text}:N_{\rm speech}=1:4$, and GLM-4-Voice is a 40-layer SLM with a text-speech interleaved schedule of $N_{\rm text}:N_{\rm speech}=13:26$.
We use greedy decoding for GLM-4-Voice in our main experiments. 
For Step-Audio-2, we use nucleus sampling with \((\mathrm{temperature}=0.7, \mathrm{top\_p}=0.9)\).
These different generation configuration follows the default generation configuration of the two SLMs respectively.

\noindent\textbf{Layer-wise LM Head Training.}
We use \textit{VoiceAssistant-400K}~\cite{xie2024mini}, a spoken dialogue dataset, for layer-specific LM head training. 
We sample a subset of 18K instances and obtain pseudo-label distributions by running full-depth autoregressive generation on SLMs and train using the procedure described in Sec.~\ref{ssec:tuned_lens}.

\noindent\textbf{Evaluation.}\quad 
We follow the common SLM evaluation settings~\cite{zeng2024glm, ding2025kimi, chiang2025stitch, li2025baichuan}, evaluating on four English datasets ranging from dialogue to factual knowledge question answering (QA): AlpacaEval~\cite{alpaca_eval}, Llama Questions~\cite{nachmani2023spoken}, TriviaQA~\cite{joshi2017triviaqa}, and WebQuestion~\cite{berant2013semantic}.

\noindent\textbf{Metrics.}\quad 
For the three QA datasets, we report the accuracy against the ground truth answer using GPT-4o-mini, following~\cite{ding2025kimi}.
For AlpacaEval, we use LLM-as-a-Judge to evaluate the response and normalize the score to 0 to 100~\cite{chiang-lee-2023-large}.
When reporting the performance on the above four datasets, we evaluate the ASR transcription of speech output from the interleaved SLM.\footnote{We synthesize the audio from the speech tokens, run ASR (\textit{Whisper-large-v3}), and evaluate the transcription. We also evaluate performance using text tokens and find a similar performance trend and even much better results.} 

% To quantify early exit efficiency, we log the exit layer at each decoding step and report the average early exit layer for text and speech positions, and the depth reduction (layer saving) percentage (\%).
To quantify early-exit efficiency, we log the exit layer $\ell_t$ at every decoding step $t$ and report the average exit depth $\bar\ell=\frac{1}{|T|}\sum_{t\in T}\ell_t$ for text and speech positions respectively, together with the resulting layer saving
\begin{equation}
\mathrm{LayerSaving}
= \Big(1-\tfrac{\bar\ell}{L}\Big)\times 100\%
= \mathbb{E}_{t}\!\left[\frac{L-\ell_t}{L}\right]\times 100\%. 
% \frac{1}{|\mathcal{S}|}\sum_{t\in\mathcal{S}}\frac{L-\ell_t}{L}\times 100\%.
\label{eq:saving}
\end{equation}
The reported saving reflects the exact exit layer taken at each position for any model and interleaving ratio. %We compute this directly from the logged per-step depths, so that the reported saving reflects the exact exit layer taken at each position for any model and interleaving ratio.
We report layer saving as a hardware-agnostic measure of the compute SPAR-K saves, rather than wall-clock latency. 
Wall-clock latency is not determined solely by arithmetic computation in transformer layers, which is what SPAR-K reduces; it also includes substantial software overhead~\cite{vellaisamy2025characterizing} independent of model depth, such as time spent dispatching operations and launching GPU kernels at each decoding step. 
This large unoptimized overhead is not the focus of this work, and it dilutes the measured speedup SPAR-K actually achieves. %is especially large in an unoptimized decoding loop like ours, whose optimization is not the focus of this work, and it dilutes the measured speedup SPAR-K actually achieves. 
Layer saving, by contrast, is invariant to this overhead and serves as both a faithful and a conservative proxy for the achievable gain.

Finally, to evaluate the speech output quality, we further report the following three metrics: (i) \textbf{objective MOS} evaluated by UTMOS-v2~\cite{baba2024t05}; (ii) \textbf{text--speech alignment} between the interleaved text and the speech tokens using ASR-WER; and (iii) \textbf{subjective human MOS on speech naturalness}, collected via Amazon Mechanical Turk under the ITU-T P.800.1 Absolute Category Rating protocol.
More specifically, for text-speech alignment, we compute \textbf{ASR-WER} by transcribing the synthesized speech with \textit{Whisper-large-v3}~\cite{radford2023robust} and measuring word error rate against the text tokens generated by the SLM; for human MOS evaluations, we randomly select 25 samples from each dataset, shuffle the resulting 100 samples, and ask annotators to rate the naturalness of each sample on a 1--5 scale, collecting at least 3 valid ratings per sample after quality filtering to compute a reliable mean MOS for each method under each SLM.

\begin{comment}
To evaluate the speech output quality, we further report the following two metrics: 
(i) \textbf{MOS} evaluated by UTMOS-v2~\cite{baba2024t05}, and (ii) \textbf{text--speech alignment} between the interleaved text and the speech tokens using word error rate (WER). 
For speech-text alignment, we compute \textbf{ASR-WER} by transcribing the synthesized speech with \textit{Whisper-large-v3}~\cite{radford2023robust} and measuring word error rate against the text tokens generated by the SLM.
Finally, to quantify early exit \textbf{efficiency}, we log the exit layer at each decoding step and report the average early exit layer for text and speech positions, and speedup percentage (\%). 
\end{comment}

\begin{table*}[t]
\caption{Speech Transcription Performance across three QA datasets in Accuracy (\%) and AlpacaEval in LLM-judge score (0-100); speech quality in MOS, Human-MOS on speech naturalness (H-MOS) and ASR-WER; early exit efficiency displayed as text/speech average exit layer (Exit Layer) and layer saving (speedup) percentage (\%).
For the baselines, the row indexing corresponds to the numbering in Section~\ref{subsection: Compared Methods}.
The early exit configuration is denoted using the early exit method (the even, odd, or triple schedule in SPAR-K or Conf for confidence-based early exit), and the number in the parentheses denote the layer of early exit. 
For the confidence-based early exit, the early exit layer is determined by the entropy, and the layer in the parenthesis is the layer where we start to apply the entropy threshold and decide whether to early exit.} 
\begin{center}
% \centering
\small
\setlength{\tabcolsep}{3.4pt} %3.5pt}
\renewcommand{\arraystretch}{1.05}

\newcommand{\baselinerow}[1]{
  % \addlinespace[2pt]
  \multicolumn{12}{c}{\textbf{#1}}\\
  % \addlinespace[2pt]
}

\newcommand{\sectionrow}[1]{
  % \addlinespace[2pt]
  \multicolumn{12}{c}{\textit{#1}}\\
  \addlinespace[2pt]
}
\newcommand{\best}[1]{\textbf{#1}}

\begin{adjustbox}{max width=\textwidth,center}
\begin{tabular}{% l 
  c 
  c
  c
  @{\hspace{6pt}}|@{\hspace{6pt}}
  *{5}{S[table-format=2.2]}
  @{\hspace{6pt}}|@{\hspace{6pt}}
  S[table-format=1.3]
  S[table-format=1.3]
  S[table-format=2.2]
  @{\hspace{6pt}}|@{\hspace{6pt}}
  c
  c
} % , table-space-text-post=\%] %S[table-format=1.4] c
\toprule % \multirow{2}{*}{Model} &
\multicolumn{3}{c}{Configurations} &
\multicolumn{5}{c}{Transcription Performance} &
\multicolumn{3}{c}{Speech Quality} &
\multicolumn{2}{c}{Efficiency: Exit Layer} \\ % Avg. 
\cmidrule{1-3}\cmidrule{4-8}\cmidrule{9-11}\cmidrule{12-13} 

Row ID & Text EE & Speech EE &
{AlpacaE} & {LlamaQA} & {TQA} & {WebQA} & {\textit{Mean}} &
{MOS $\uparrow$} & {H-MOS $\uparrow$} & {ASR-WER $\downarrow$} & 
{Text} & {Speech (LayerSaving\%)} \\ %Avg. Exit Layer T/S} \\ % \ell_{EE}$ T/S $\downarrow$} \\ %(T/S)} \\& Text EE
\midrule
\sectionrow{\textbf{\underline{Step-Audio-2}}}
\sectionrow{Baselines} 
S1 & -- & -- & 45.98 & 68.00 & 49.10 & 53.80 & 54.22 & 3.710 & 3.998 & 1.51 & 28 & 28 \\
% greedy & disable & 54.07 & 72.33 & 54.20 & 59.40 & 2.3580 & 4.43 & 28 \\
S2 & -- & Fixed (25) & 35.58 & 67.33 & 46.70 & 50.90 & 50.13 & 3.058 & 3.810 & 3.40 & 28 & 25.00 (11\%$\uparrow$) \\
% (0.7, 0.9) & $\text{S - }$0.3 (26) & 57.14 & 74.00 & 50.60 & 55.70 & 1.6353 & 10.47 & 28 / 26.05 \\
S3 & -- & Conf (26) & 24.92 & 65.00 & 28.50 & 47.60 & 41.51 & 1.651 & 2.807 & 11.01 & 28 & 26.07 (6.9\%$\uparrow$)  \\

\sectionrow{SPAR-K (Ours)} 
\rowcolor{gray!15} %gray!15} \best{BEST} & 
S4 & -- & $\text{Even}$ (22) & 45.28 & 63.00 & 49.20 & 55.40 & 53.22 & 3.643 & 3.940 & 1.58 & 28 & 25.00 \textcolor{red!60!black}{(11\%$\uparrow$)} \\ 
\rowcolor{gray!15}
S5 & -- & $\text{Odd}$ (22) & 45.98 & 67.00 & 48.70 & 53.30 & 53.75 & 3.650 & 4.133 & 2.19 & 28 & 25.00 \textcolor{red!60!black}{(11\%$\uparrow$)} \\
\rowcolor{gray!15}
S6 & -- & $\text{Triple}$ (22) & 46.33 & 67.33 & 51.00 & 56.50 & 55.29 & 3.668 & 3.977 & 1.51 & 28 & 25.00 \textcolor{red!60!black}{(11\%$\uparrow$)} \\ 

\sectionrow{\textit{Analyses}}

\rowcolor{gray!15}
S7 & -- & $\text{Even}$ (21) & 44.37 & 66.33 & 49.10 & 54.00 & 53.45 & 3.652 & {--} & 1.57 & 28 & 24.50 \textcolor{red!60!black}{(12\%$\uparrow$)} \\ 
%44.37%	66.33%	49.10%	54.00%	53.45%	3.6516	1.57%

\midrule
\midrule
\sectionrow{\textbf{\underline{GLM-4-Voice}}}
% -------------------- (Rows from your sheet) --------------------
\sectionrow{Baselines} 
 G1 & -- & -- & 51.06 & 61.33 & 44.90 & 52.20 & 52.37 & 2.982 & 3.345 & 4.31 & 40 & 40 \\

G2 & -- & Fixed (38) & 44.17 & 65.33 & 45.00 & 51.40 & 51.48 & 2.662 & 2.083 & 43.60 & 40 & 38.00 (5.0\%$\uparrow$) \\

G3 & -- & Conf (36) & 50.20 & 64.67 & 45.70 & 51.00 & 52.89 & 2.866 & 3.060 & 7.62 & 40 & 37.03 (7.4\%$\uparrow$)  \\

\sectionrow{SPAR-K (Ours)}

\rowcolor{gray!15} %\best{BEST} & 
G4 & -- & $\text{Even}$ (36) & 50.70 & 60.00 & 42.00 & 50.60 & 50.83 & 2.950 & 3.345 & 5.36 & 40 & 38.00 \textcolor{red!60!black}{(5.0\%$\uparrow$)} \\
\rowcolor{gray!15}
G5 & -- & $\text{Odd}$ (36) & 49.75 & 58.33 & 43.00 & 50.30 & 50.35 & 2.982 & 3.187 & 5.06 & 40 & 38.00 \textcolor{red!60!black}{(5.0\%$\uparrow$)}\\
\rowcolor{gray!15}
G6 & -- & $\text{Triple}$ (37) & 47.84 & 63.33 & 44.70 & 51.90 & 51.94 & 2.909 & 3.182 & 5.77 & 40 & 38.00 \textcolor{red!60!black}{(5.0\%$\uparrow$)}\\

\sectionrow{\textit{Analyses}}

\rowcolor{gray!15}
G7 & -- & $\text{Even}$ (31) & 47.69 & 58.33 & 38.00 & 44.80 & 47.21 & 2.960 & {--} & 8.85 & 40 & 35.50 \textcolor{red!60!black}{(11\%$\uparrow$)} \\
\rowcolor{gray!15}
G8 & -- & $\text{Even}$ (33) & 48.19 & 57.33 & 41.80 & 48.50 & 48.96 & 2.981 & {--} & 7.13 & 40 & 36.50 \textcolor{red!60!black}{(8.8\%$\uparrow$)} \\
\rowcolor{gray!15}
G9 & -- & $\text{Even}$ (35) & 49.45 & 61.00 & 41.20 & 49.50 & 50.29 & 2.942 & {--} & 6.44 & 40 & 37.00 \textcolor{red!60!black}{(7.5\%$\uparrow$)} \\
G10 & $\text{Even}$ (36) & -- & 10.00 & 42.00 & 0.00 & 21.70 & 18.43 & 2.979 & {--} & 12.80 & 38.00 & 40 \\
% Greedy & $\text{S - }\mathrm{2+}1_{\text{even}}$(37) & 56.73 & 75.00 & 51.00 & 55.10 & 2.9089 & 5.77 & 40 / 38 \\
G11 & Conf (36) & $\text{Even}$ (36) & 48.64 & 59.00 & 38.50 & 48.30 & 48.61 & 2.949 & {--} & 5.15 &  38.02 & 38.00 (5.0\%$\uparrow$) \\

\bottomrule
\end{tabular}
\end{adjustbox}
\vspace{-6pt}
\label{tab:main_results_grouped}
\end{center}
\end{table*}

\subsection{Compared Methods}
\label{subsection: Compared Methods}
We include the following four methods for comparison.
Unless otherwise specified, all the methods with early exit \textbf{only early exit the speech tokens}.
(1) \textbf{No early exit}: using full-depth transformer layers to generate the next token.
(2) \textbf{Fixed-layer early exit}: always exiting on a fixed layer for the speech tokens.
(3) \textbf{Confidence-based early exit}: Given a hidden representation $h_t^{\ell}$, calculate the next token prediction at that layer and obtain the entropy of the distribution. 
Next, compare the entropy with a predefined threshold and early exit only if the entropy is lower than the threshold.
This is a widely used method in text-only LLMs.
We carefully tune the threshold to obtain the best performance.
(4) \textbf{SPAR-K}: Our proposed method. 
We report the performance of three difference schedules, and the early exit layer is selected based on the ASR-WER and MOS using a held-out set from VoiceAssistant-400K.

\subsection{Main Results}
\noindent\textbf{Overall performance.} 
SPAR-K largely preserves SLM performance on average for the four evaluated datasets, with no accuracy drop on Step-Audio-2 and a maximum average accuracy drop of 0.82\% on GLM-4-Voice (Table~\ref{tab:main_results_grouped}: S6, G6).
For Step-Audio-2, in the best-performing configuration, $\textit{Triple}  (22)$ schedule, with notably 11\% layer reduction in speech tokens, maintains ASR-WER (no increase) while only slightly reducing MOS ($3.710\rightarrow3.668$, -1.12\%).
For GLM-4-Voice, the best overall balance of GLM-4-Voice is achieved by $\textit{Even}  (36)$: speech tokens exit at an average depth of 38 (5\% layer reduction on speech decoding) without introducing additional compute, and with only modest speech-quality changes (MOS $2.982\rightarrow2.950$, 1.07\% drop; ASR-WER $4.31\%\rightarrow5.36\%$).

\noindent\textbf{Fixed-Layer EE Causes Severe Speech Degradation.}
As discussed in Sec.~\ref{ssec:motivation}, a naive \textit{Fixed-Layer} early exit policy is not feasible for SLM speech decoding. 
This is corroborated by the quantitative results in row S2, G2. Both exhibit pronounced drops in speech quality. Possibly due to early exit distribution shift~\cite{bae2023fast}, we find that after the model finishes generating the text-related context, it continues producing redundant speech tokens without properly terminating, resulting in high WER. 

\begin{comment}
\begin{table*}[t]
\centering
\caption{Human MOS evaluation (ITU-T P.800.1 ACR protocol). Mean MOS and standard deviation per method; row IDs match Table~\ref{tab:main_results_grouped}.}
\label{tab:mos_human}
\small % bigger font
\renewcommand{\arraystretch}{1.1}       % taller rows
\setlength{\tabcolsep}{8pt}           % wider column gaps
\begin{tabular}{l ccc >{\columncolor{gray!20}}c >{\columncolor{gray!20}}c >{\columncolor{gray!20}}c ccc >{\columncolor{gray!20}}c >{\columncolor{gray!20}}c >{\columncolor{gray!20}}c} %ccc cccccc}
\toprule
Model & \multicolumn{6}{c}{Step-Audio-2} & \multicolumn{6}{c}{GLM-4-Voice} \\
\cmidrule(lr){2-7} \cmidrule(lr){8-13}
Row ID & S1 & S2 & S3 & S4 & S5 & S6 & G1 & G2 & G3 & G4 & G5 & G6 \\
\midrule
Mean MOS $\uparrow$ & 3.998 & 3.810 & 2.807 & 3.940 & 4.133 & 3.977 & 3.345 & 2.083 & 3.060 & 3.345 & 3.187 & 3.182 \\
Std  & 0.518 & 0.640 & 0.820 & 0.470 & 0.552 & 0.560 & 0.557 & 0.723 & 0.693 & 0.591 & 0.591 & 0.582 \\
\bottomrule
\end{tabular}
% \vspace{-5pt}
\end{table*}
\end{comment}

\noindent\textbf{SPAR-K vs.\ Confidence-Based Early Exit.} 
\textit{Confidence-based early exit} is a common paradigm in LLM inference.
When applying confidence-based early exit to SLMs for the speech tokens, we find that it leads to a significant performance (quality) drop for Step-Audio-2 (row S3).
For GLM-4-Voice, confidence-based early exit can work under carefully tuned entropy thresholds (row G3).
This shows that confidence-based early exit is sensitive to the SLM backbones.
Moreover, a significant drawback of confidence-based early exit is that it typically requires additional computation: to decide the entropy, we need to project the hidden representation to the vocabulary space; if the token prediction from the intermediate layer isn't used due to insufficient confidence, the computation of calculating the token distribution at the intermediate layer is a waste.
In contrast, since \textbf{SPAR-K} is based on a fixed schedule, we will not waste computation on calculating token distribution that may not be usable.

\noindent\textbf{Discussion over schedule variants.}
We first clarify why different schedules can attain the same efficiency. 
By Eq.~\eqref{eq:saving}, the average exit depth of a speech chunk is the mean of its per-position exit layers, which depends on \textit{how many} positions run at full depth $L$ versus $\ell_{EE}$, not on their ordering. 
Two schedules with the same number of full-depth positions per chunk therefore yield identical layer savings; the schedule acts not as an \textit{efficiency} control but as a \textit{quality} control, governing how early-exit positions are spaced between full-depth ``refresh'' steps. 
Table~\ref{tab:main_results_grouped} shows this: for Step-Audio-2, whose interleaving cycle has only four speech positions, the \textit{Even}, \textit{Odd}, and \textit{Triple} schedules each use two full-depth and two early-exit positions per chunk, sharing an identical $25.0$ average exit depth (S4--S6); on GLM-4-Voice we likewise match savings across schedules (G4--G6), so the comparison isolates spacing effects.

\noindent Under this matched-efficiency view, we find that Step-Audio-2 tolerates a larger period $K$: the \textit{Triple} schedule performs comparably to \textit{Even} and \textit{Odd} (S4--S6). 
On GLM-4-Voice, by contrast, \textit{Triple}(37) yields slightly lower MOS than \textit{Even}(36). 
We attribute this to GLM-4-Voice's much longer speech block (26 speech positions per cycle versus 4). 
As $K$ grows, the spacing between refreshes widens, so more consecutive tokens are decoded shallowly and error can accumulate over a longer stretch -- an effect amplified by GLM-4-Voice's long chunk. 
Overall, however, the schedules do not differ significantly, indicating that SPAR-K is robust to this choice.

\subsection{Human Speech Naturalness Evaluation: Human MOS result matches UTMOS-v2 result}
To verify that early-exit mechanisms preserve speech output quality, we complement our objective UTMOS-v2 assessment with a human evaluation of speech naturalness for audio generated by SPAR-K and the baseline methods. % We randomly select 25 samples from each evaluation dataset (Table~\ref{tab:main_results_grouped}), shuffle the resulting 100 samples, and ask annotators to rate the naturalness of each sample on a 1–5 scale, collecting at least 3 valid ratings per sample after quality filtering to compute a reliable mean MOS for each method under each base model. 

As shown in Table~\ref{tab:main_results_grouped} (H-MOS column), importantly, SPAR-K (rows S4–S6 and G4–G6) matches the naturalness of the original non-early-exit models (row S1 and G1). Moreover, the relative MOS ordering across SPAR-K and other compared methods (Fixed-layer: S2, G2; Confidence: S3, G3) is consistent with the UTMOS-v2 results (Table~\ref{tab:main_results_grouped} - Speech Quality MOS column).

\subsection{Analyses}
\noindent\textbf{Early Exit Layer Analyses of SPAR-K} \quad
We investigate the sensitivity of the SPAR-K policy to the early exit layer hyperparameter, $\ell_{EE}$. 
As shown in row S7 and G7-G9, as $\ell_{EE}$ decreases, we observe a gradual increase in ASR-WER, rather than an abrupt collapse in performance. 
This trend suggests that perceptual quality can be preserved within a specific depth range.

\noindent\textbf{Exploring Text Early Exit Policies} \quad
We also try to apply SPAR-K on text tokens.
However, as shown in row G10, we find a severe drop in transcription performance.
This suggests that text tokens require finer-grained early exit control to dynamically allocate decoding depth per token. %(Table~\ref{tab:main_results_grouped})
We also try to apply confidence-based early exit on text tokens and apply SPAR-K on the speech tokens.
The result is shown in row G11. % (Table~\ref{tab:main_results_grouped}).
With carefully chosen confidence thresholds, the average performance drop is reduced to 7.18\%, while the speech quality remains.
This highlights that the distinct properties in text and speech tokens require different early exit policy.

\section{Implementation Details}
\subsection{Setup of Confidence-based Early Exit Experiments}
As mentioned in Section~\ref{subsection: Compared Methods}, in \textbf{Confidence-based early exit} experiments, we carefully tune the entropy threshold to obtain the best performance for the method. After large-scale experiments, the final entropy threshold for GLM-4-Voice is set to 0.5 (Table.~\ref{tab:main_results_grouped}: G3) and the entropy threshold for Step-Audio-2 is set to 0.1 (Table.~\ref{tab:main_results_grouped}: S3).

% \vspace{-3pt}
\section{Limitations}
We acknowledge some limitations of SPAR-K. 
First, our evaluation focuses on two SLMs in the interleaved paradigm, where a single autoregressive stream alternates text and speech tokens. 
Other SLM architectures like Kimi-Audio~\cite{ding2025kimi} generate text and speech tokens through separate parallel heads on top of a shared transformer backbone. % Kimi-Audio~\cite{ding2025kimi}. %, in which text and audio tokens are produced simultaneously by two dedicated heads at each step. 
The speech alternating-depth schedule is in principle applicable to the audio-head stream in such models, but we have not empirically validated SPAR-K in this setting. %; the interaction between the shared backbone and the two parallel heads may require additional design consideration. 
Second, the early-exit layer $\ell_{EE}$ is currently selected via held-out validation; adaptive or learned scheduling policies with no extra compute that allocate depth based on model backbone could plausibly yield further gains. 
Third, SPAR-K accelerates only speech positions; our analyses in Section~V-E (rows G9--G10) suggest that text tokens require finer-grained policies, and combining modality-specific schedules across both token types is a promising direction.

\begin{comment}
Several limitations of SPAR-K motivate future investigation.
First, our evaluation focuses on two interleaved SLMs with similar decoder-only LM backbones; whether the speech alternating-depth schedule transfers to other SLM paradigms, such as multi-stream codec-LMs that model parallel acoustic streams or RQ-Transformer-based hierarchical SLMs, remains to be examined. % parallel head
Second, the period $K$ and the early-exit layer $\ell_{EE}$ are currently selected via held-out validation; adaptive or learned scheduling policies that allocate depth based on input context could plausibly yield further gains while preserving the no-extra-compute property.
Lastly, SPAR-K targets only speech positions: our analyses in Section~V-E (rows G9--G10) suggest that text tokens require finer-grained policies, and combining modality-specific schedules is a promising direction.
\end{comment}

% \vspace{-3pt}
\section{Conclusion}
We present SPAR-K, a modality-aware early exit framework that accelerates interleaved spoken language models without retraining. 
Motivated by the local predictability of speech tokens, SPAR-K replaces confidence-based per-token exit with a scheduled periodic alternating depth pattern: most speech positions exit at a fixed intermediate layer, while periodic full-depth refresh steps mitigate distribution shift.
On Step-Audio-2-mini and GLM-4-Voice across reasoning, factual QA, and dialogue benchmarks, SPAR-K reduces average speech decoding depth by up to 11\% with at most a 0.82\% drop in question-answering accuracy, and preserves perceptual quality across both objective and human listening evaluations. 
Our analysis further shows that confidence-based exit criteria, while effective for text LLMs, are suboptimal for interleaved SLMs, establishing modality-aware depth control as a promising and complementary direction for SLM inference acceleration.

\bibliographystyle{IEEEtran}
\bibliography{refs}

@article{arora2025landscape,
  title={On the landscape of spoken language models: A comprehensive survey},
  author={Arora, Siddhant and Chang, Kai-Wei and Chien, Chung-Ming and Peng, Yifan and Wu, Haibin and Adi, Yossi and Dupoux, Emmanuel and Lee, Hung-Yi and Livescu, Karen and Watanabe, Shinji},
  journal={arXiv preprint arXiv:2504.08528},
  year={2025}
}

@article{schuster2022confident,
  title={Confident adaptive language modeling},
  author={Schuster, Tal and Fisch, Adam and Gupta, Jai and Dehghani, Mostafa and Bahri, Dara and Tran, Vinh and Tay, Yi and Metzler, Donald},
  journal={Advances in Neural Information Processing Systems},
  volume={35},
  pages={17456--17472},
  year={2022}
}

@inproceedings{elhoushi2024layerskip,
  title={Layerskip: Enabling early exit inference and self-speculative decoding},
  author={Elhoushi, Mostafa and Shrivastava, Akshat and Liskovich, Diana and Hosmer, Basil and Wasti, Bram and Lai, Liangzhen and Mahmoud, Anas and Acun, Bilge and Agarwal, Saurabh and Roman, Ahmed and others},
  booktitle={Proceedings of the 62nd Annual Meeting of the Association for Computational Linguistics (Volume 1: Long Papers)},
  pages={12622--12642},
  year={2024}
}

@article{belrose2023eliciting,
  title={Eliciting latent predictions from transformers with the tuned lens},
  author={Belrose, Nora and Furman, Zach and Smith, Logan and Halawi, Danny and Ostrovsky, Igor and McKinney, Lev and Biderman, Stella and Steinhardt, Jacob},
  journal={arXiv preprint arXiv:2303.08112},
  year={2023}
}

@article{xie2024mini,
  title={Mini-omni: Language models can hear, talk while thinking in streaming},
  author={Xie, Zhifei and Wu, Changqiao},
  journal={arXiv preprint arXiv:2408.16725},
  year={2024}
}

@article{zeng2024glm,
  title={Glm-4-voice: Towards intelligent and human-like end-to-end spoken chatbot},
  author={Zeng, Aohan and Du, Zhengxiao and Liu, Mingdao and Wang, Kedong and Jiang, Shengmin and Zhao, Lei and Dong, Yuxiao and Tang, Jie},
  journal={arXiv preprint arXiv:2412.02612},
  year={2024}
}

@article{wu2025step,
  title={Step-audio 2 technical report},
  author={Wu, Boyong and Yan, Chao and Hu, Chen and Yi, Cheng and Feng, Chengli and Tian, Fei and Shen, Feiyu and Yu, Gang and Zhang, Haoyang and Li, Jingbei and others},
  journal={arXiv preprint arXiv:2507.16632},
  year={2025}
}

@article{ding2025kimi,
  title={Kimi-audio technical report},
  author={Ding, Ding and Ju, Zeqian and Leng, Yichong and Liu, Songxiang and Liu, Tong and Shang, Zeyu and Shen, Kai and Song, Wei and Tan, Xu and Tang, Heyi and others},
  journal={arXiv preprint arXiv:2504.18425},
  year={2025}
}

@article{chiang2025stitch,
  title={Stitch: Simultaneous thinking and talking with chunked reasoning for spoken language models},
  author={Chiang, Cheng-Han and Wang, Xiaofei and Li, Linjie and Lin, Chung-Ching and Lin, Kevin and Liu, Shujie and Wang, Zhendong and Yang, Zhengyuan and Lee, Hung-yi and Wang, Lijuan},
  journal={arXiv preprint arXiv:2507.15375},
  year={2025}
}

@article{li2025baichuan,
  title={Baichuan-audio: A unified framework for end-to-end speech interaction},
  author={Li, Tianpeng and Liu, Jun and Zhang, Tao and Fang, Yuanbo and Pan, Da and Wang, Mingrui and Liang, Zheng and Li, Zehuan and Lin, Mingan and Dong, Guosheng and others},
  journal={arXiv preprint arXiv:2502.17239},
  year={2025}
}

@article{nachmani2023spoken,
  title={Spoken question answering and speech continuation using spectrogram-powered llm},
  author={Nachmani, Eliya and Levkovitch, Alon and Hirsch, Roy and Salazar, Julian and Asawaroengchai, Chulayuth and Mariooryad, Soroosh and Rivlin, Ehud and Skerry-Ryan, RJ and Ramanovich, Michelle Tadmor},
  journal={arXiv preprint arXiv:2305.15255},
  year={2023}
}

@inproceedings{joshi2017triviaqa,
  title={Triviaqa: A large scale distantly supervised challenge dataset for reading comprehension},
  author={Joshi, Mandar and Choi, Eunsol and Weld, Daniel S and Zettlemoyer, Luke},
  booktitle={Proceedings of the 55th Annual Meeting of the Association for Computational Linguistics (Volume 1: Long Papers)},
  pages={1601--1611},
  year={2017}
}

@inproceedings{berant2013semantic,
  title={Semantic parsing on freebase from question-answer pairs},
  author={Berant, Jonathan and Chou, Andrew and Frostig, Roy and Liang, Percy},
  booktitle={Proceedings of the 2013 conference on empirical methods in natural language processing},
  pages={1533--1544},
  year={2013}
}

@misc{alpaca_eval,
  author = {Xuechen Li and Tianyi Zhang and Yann Dubois and Rohan Taori and Ishaan Gulrajani and Carlos Guestrin and Percy Liang and Tatsunori B. Hashimoto },
  title = {AlpacaEval: An Automatic Evaluator of Instruction-following Models},
  year = {2023},
  publisher = {GitHub},
  journal = {GitHub repository},
  howpublished = {\url{https://github.com/tatsu-lab/alpaca\_eval}}
}

@inproceedings{baba2024t05,
  title={The t05 system for the voicemos challenge 2024: Transfer learning from deep image classifier to naturalness mos prediction of high-quality synthetic speech},
  author={Baba, Kaito and Nakata, Wataru and Saito, Yuki and Saruwatari, Hiroshi},
  booktitle={2024 IEEE Spoken Language Technology Workshop (SLT)},
  pages={818--824},
  year={2024},
  organization={IEEE}
}

@inproceedings{radford2023robust,
  title={Robust speech recognition via large-scale weak supervision},
  author={Radford, Alec and Kim, Jong Wook and Xu, Tao and Brockman, Greg and McLeavey, Christine and Sutskever, Ilya},
  booktitle={International conference on machine learning},
  pages={28492--28518},
  year={2023},
  organization={PMLR}
}

@inproceedings{chou2023toward,
  title={Toward joint language modeling for speech units and text},
  author={Chou, Ju-Chieh and Chien, Chung-Ming and Hsu, Wei-Ning and Livescu, Karen and Babu, Arun and Conneau, Alexis and Baevski, Alexei and Auli, Michael},
  booktitle={Findings of the Association for Computational Linguistics: EMNLP 2023},
  pages={6582--6593},
  year={2023}
}

@article{zeng2024scaling,
  title={Scaling speech-text pre-training with synthetic interleaved data},
  author={Zeng, Aohan and Du, Zhengxiao and Liu, Mingdao and Zhang, Lei and Jiang, Shengmin and Dong, Yuxiao and Tang, Jie},
  journal={arXiv preprint arXiv:2411.17607},
  year={2024}
}

@inproceedings{zhang2023speechgpt,
  title={Speechgpt: Empowering large language models with intrinsic cross-modal conversational abilities},
  author={Zhang, Dong and Li, Shimin and Zhang, Xin and Zhan, Jun and Wang, Pengyu and Zhou, Yaqian and Qiu, Xipeng},
  booktitle={Findings of the Association for Computational Linguistics: EMNLP 2023},
  pages={15757--15773},
  year={2023}
}

@inproceedings{liu2025speech,
  title={Speech token prediction via compressed-to-fine language modeling for speech generation},
  author={Liu, Wenrui and Chen, Qian and Wang, Wen and Yang, Guanrou and Li, Weiqin and Fang, Minghui and Zuo, Jialong and Yang, Xiaoda and Jin, Tao and Xu, Jin and others},
  booktitle={Proceedings of the 33rd ACM International Conference on Multimedia},
  pages={10632--10641},
  year={2025}
}

@article{yang2024interleaved,
  title={Interleaved Speech-Text Language Models for Simple Streaming Text-to-Speech Synthesis},
  author={Yang, Yifan and Liu, Shujie and Li, Jinyu and Wang, Hui and Meng, Lingwei and Sun, Haiyang and Liang, Yuzhe and Ma, Ziyang and Hu, Yuxuan and Zhao, Rui and others},
  journal={arXiv preprint arXiv:2412.16102},
  year={2024}
}

@article{aylett2006language,
  title={Language redundancy predicts syllabic duration and the spectral characteristics of vocalic syllable nuclei},
  author={Aylett, Matthew and Turk, Alice},
  journal={The Journal of the Acoustical Society of America},
  volume={119},
  number={5},
  pages={3048--3058},
  year={2006},
  publisher={AIP Publishing}
}

@article{zhang2023speechtokenizer,
  title={Speechtokenizer: Unified speech tokenizer for speech large language models},
  author={Zhang, Xin and Zhang, Dong and Li, Shimin and Zhou, Yaqian and Qiu, Xipeng},
  journal={arXiv preprint arXiv:2308.16692},
  year={2023}
}

@inproceedings{leviathan2023fast,
  title={Fast inference from transformers via speculative decoding},
  author={Leviathan, Yaniv and Kalman, Matan and Matias, Yossi},
  booktitle={International Conference on Machine Learning},
  pages={19274--19286},
  year={2023},
  organization={PMLR}
}

@article{elbayad2019depth,
  title={Depth-adaptive transformer},
  author={Elbayad, Maha and Gu, Jiatao and Grave, Edouard and Auli, Michael},
  journal={arXiv preprint arXiv:1910.10073},
  year={2019}
}

@article{chen2023accelerating,
  title={Accelerating large language model decoding with speculative sampling},
  author={Chen, Charlie and Borgeaud, Sebastian and Irving, Geoffrey and Lespiau, Jean-Baptiste and Sifre, Laurent and Jumper, John},
  journal={arXiv preprint arXiv:2302.01318},
  year={2023}
}

@article{du2024cosyvoice2,
  title={Cosyvoice 2: Scalable streaming speech synthesis with large language models},
  author={Du, Zhihao and Wang, Yuxuan and Chen, Qian and Shi, Xian and Lv, Xiang and Zhao, Tianyu and Gao, Zhifu and Yang, Yexin and Gao, Changfeng and Wang, Hui and others},
  journal={arXiv preprint arXiv:2412.10117},
  year={2024}
}

@article{du2024cosyvoice,
  title={Cosyvoice: A scalable multilingual zero-shot text-to-speech synthesizer based on supervised semantic tokens},
  author={Du, Zhihao and Chen, Qian and Zhang, Shiliang and Hu, Kai and Lu, Heng and Yang, Yexin and Hu, Hangrui and Zheng, Siqi and Gu, Yue and Ma, Ziyang and others},
  journal={arXiv preprint arXiv:2407.05407},
  year={2024}
}

@article{kong2020hifi,
  title={Hifi-gan: Generative adversarial networks for efficient and high fidelity speech synthesis},
  author={Kong, Jungil and Kim, Jaehyeon and Bae, Jaekyoung},
  journal={Advances in neural information processing systems},
  volume={33},
  pages={17022--17033},
  year={2020}
}

@article{chen2023ee,
  title={Ee-llm: Large-scale training and inference of early-exit large language models with 3d parallelism},
  author={Chen, Yanxi and Pan, Xuchen and Li, Yaliang and Ding, Bolin and Zhou, Jingren},
  journal={arXiv preprint arXiv:2312.04916},
  year={2023}
}

@article{bajpai2025survey,
  title={A survey of early exit deep neural networks in nlp},
  author={Bajpai, Divya Jyoti and Hanawal, Manjesh Kumar},
  journal={arXiv preprint arXiv:2501.07670},
  year={2025}
}

@inproceedings{bae2023fast,
  title={Fast and robust early-exiting framework for autoregressive language models with synchronized parallel decoding},
  author={Bae, Sangmin and Ko, Jongwoo and Song, Hwanjun and Yun, Se-Young},
  booktitle={Proceedings of the 2023 Conference on Empirical Methods in Natural Language Processing},
  pages={5910--5924},
  year={2023}
}

@article{valade2024accelerating,
  title={Accelerating large language model inference with self-supervised early exits},
  author={Valade, Florian},
  journal={arXiv preprint arXiv:2407.21082},
  year={2024}
}

@article{zheng2025hybrid,
  title={A Hybrid Early-Exit Algorithm for Large Language Models Based on Space Alignment Decoding (SPADE)},
  author={Zheng, Bowen and Ma, Ming and Lin, Zhongqiao and Yang, Tianming},
  journal={arXiv preprint arXiv:2507.17618},
  year={2025}
}

@article{miao2024efficient,
  title={An efficient inference framework for early-exit large language models},
  author={Miao, Ruijie and Yan, Yihan and Yao, Xinshuo and Yang, Tong},
  journal={arXiv preprint arXiv:2407.20272},
  year={2024}
}

@inproceedings{cui2025recent,
  title={Recent advances in speech language models: A survey},
  author={Cui, Wenqian and Yu, Dianzhi and Jiao, Xiaoqi and Meng, Ziqiao and Zhang, Guangyan and Wang, Qichao and Guo, Steven Y and King, Irwin},
  booktitle={Proceedings of the 63rd Annual Meeting of the Association for Computational Linguistics (Volume 1: Long Papers)},
  pages={13943--13970},
  year={2025}
}

@article{defossez2024moshi,
  title={Moshi: a speech-text foundation model for real-time dialogue},
  author={D{\'e}fossez, Alexandre and Mazar{\'e}, Laurent and Orsini, Manu and Royer, Am{\'e}lie and P{\'e}rez, Patrick and J{\'e}gou, Herv{\'e} and Grave, Edouard and Zeghidour, Neil},
  journal={arXiv preprint arXiv:2410.00037},
  year={2024}
}

@article{yoon2022hubert,
  title={Hubert-ee: Early exiting hubert for efficient speech recognition},
  author={Yoon, Ji Won and Woo, Beom Jun and Kim, Nam Soo},
  journal={arXiv preprint arXiv:2204.06328},
  year={2022}
}

@inproceedings{chiang-lee-2023-large,
    title = "Can Large Language Models Be an Alternative to Human Evaluations?",
    author = "Chiang, Cheng-Han  and
      Lee, Hung-yi",
    editor = "Rogers, Anna  and
      Boyd-Graber, Jordan  and
      Okazaki, Naoaki",
    booktitle = "Proceedings of the 61st Annual Meeting of the Association for Computational Linguistics (Volume 1: Long Papers)",
    month = jul,
    year = "2023",
    address = "Toronto, Canada",
    publisher = "Association for Computational Linguistics",
    url = "https://aclanthology.org/2023.acl-long.870/",
    doi = "10.18653/v1/2023.acl-long.870",
    pages = "15607--15631"
}

@article{pope2023efficiently,
  title={Efficiently scaling transformer inference},
  author={Pope, Reiner and Douglas, Sholto and Chowdhery, Aakanksha and Devlin, Jacob and Bradbury, James and Heek, Jonathan and Xiao, Kefan and Agrawal, Shivani and Dean, Jeff},
  journal={Proceedings of machine learning and systems},
  volume={5},
  pages={606--624},
  year={2023}
}

@inproceedings{wright2024training,
  title={Training early-exit architectures for automatic speech recognition: Fine-tuning pre-trained models or training from scratch},
  author={Wright, George August and Cappellazzo, Umberto and Zaiem, Salah and Raj, Desh and Yang, Lucas Ondel and Falavigna, Daniele and Ali, Mohamed Nabih and Brutti, Alessio},
  booktitle={2024 IEEE International Conference on Acoustics, Speech, and Signal Processing Workshops (ICASSPW)},
  pages={685--689},
  year={2024},
  organization={IEEE}
}

@inproceedings{lasbordes2025splitformer,
  title={Splitformer: An improved early-exit architecture for automatic speech recognition on edge devices},
  author={Lasbordes, Maxence and Falavigna, Daniele and Brutti, Alessio},
  booktitle={2025 33rd European Signal Processing Conference (EUSIPCO)},
  pages={341--345},
  year={2025},
  organization={IEEE}
}

@article{lin2024daisy,
  title={DAISY: Data adaptive self-supervised early exit for speech representation models},
  author={Lin, Tzu-Quan and Lee, Hung-yi and Tang, Hao},
  journal={arXiv preprint arXiv:2406.05464},
  year={2024}
}

@article{hassid2023textually,
  title={Textually pretrained speech language models},
  author={Hassid, Michael and Remez, Tal and Nguyen, Tu Anh and Gat, Itai and Conneau, Alexis and Kreuk, Felix and Copet, Jade and Defossez, Alexandre and Synnaeve, Gabriel and Dupoux, Emmanuel and others},
  journal={Advances in Neural Information Processing Systems},
  volume={36},
  pages={63483--63501},
  year={2023}
}

@inproceedings{chen2025slam,
  title={Slam-omni: Timbre-controllable voice interaction system with single-stage training},
  author={Chen, Wenxi and Ma, Ziyang and Yan, Ruiqi and Liang, Yuzhe and Li, Xiquan and Xu, Ruiyang and Niu, Zhikang and Zhu, Yanqiao and Yang, Yifan and Liu, Zhanxun and others},
  booktitle={Findings of the Association for Computational Linguistics: ACL 2025},
  pages={2262--2282},
  year={2025}
}

@inproceedings{wang2025vocalnet,
  title={VocalNet: Speech LLMs with Multi-Token Prediction for Faster and High-Quality Generation},
  author={Wang, Yuhao and Liu, Heyang and Cheng, Ziyang and Wu, Ronghua and Gu, Qunshan and Wang, Yanfeng and Wang, Yu},
  booktitle={Proceedings of the 2025 Conference on Empirical Methods in Natural Language Processing},
  pages={19595--19612},
  year={2025}
}

@article{lin2025accelerating,
  title={Accelerating Autoregressive Speech Synthesis Inference With Speech Speculative Decoding},
  author={Lin, Zijian and Zhang, Yang and Yuan, Yougen and Yan, Yuming and Liu, Jinjiang and Wu, Zhiyong and Hu, Pengfei and Yu, Qun},
  journal={arXiv preprint arXiv:2505.15380},
  year={2025}
}

@inproceedings{vellaisamy2025characterizing,
  title={Characterizing and optimizing llm inference workloads on cpu-gpu coupled architectures},
  author={Vellaisamy, Prabhu and Labonte, Thomas and Chakraborty, Sourav and Turner, Matt and Sury, Samantika and Shen, John Paul},
  booktitle={2025 IEEE International Symposium on Performance Analysis of Systems and Software (ISPASS)},
  pages={49--61},
  year={2025},
  organization={IEEE}
}

\section{Author Contributions}
All authors made meaningful contributions to the experimental design, the writing process, and the refinement of the paper. 
% Although each author is involved in multiple aspects of the project, their primary contributions are summarized below:
% All authors contributed meaningfully to the design of the experiments, the writing, and the refinement of the paper. 
While all authors were involved in multiple aspects of the project, their primary contributions are highlighted below:

\textbf{Hsiao-Ying Huang} leads the overall project direction. 
She conducts all the experiments, designs the analysis, and drafts most of the manuscript. 
\textbf{Cheng-Han Chiang} proposed the initial idea and participated in writing the paper.
\textbf{Cheng-Han Chiang} and \textbf{Hung-yi Lee} contributed deep technical expertise, helped shape the research direction, and provided crucial guidance and feedback on the experimental methodology and manuscript development.

\end{document}